# Generation of Heterogeneous PET Images from Uniform Organ Activity Maps Using a Pretrained Domain-Adapted Diffusion Model

***Suya Li*[1,2]*, Kaushik Dutta*[1,2]*, Debojyoti Pal*[1,2]*, Jingqin Luo*[3]*, and Kooresh I. Shoghi*[1,2,4*]**

[1]*Mallinckrodt Institute of Radiology, Washington University School of Medicine, St. Louis, USA*
[2]*Imaging Science Program, McKelvey School of Engineering, Washington University in St Louis, St. Louis, USA*
[3]*Department of Surgery, Washington University School of Medicine, St. Louis, USA*
[4]*Department of Biomedical Engineering, Washington University in St Louis, St. Louis, USA*

## Abstract

*Synthetic PET images are valuable for quantitative imaging workflow development, scalable virtual imaging trials, and deep learning model training, but conventional physics-based simulation approaches are computationally intensive, limited in anatomical variability, and often fail to capture heterogeneous PET uptake. This study developed a pretrained domain-adapted diffusion (PAD) model for anatomy-conditioned PET synthesis from uniform organ activity maps. PAD adopts a natural-image pretrained text-to-image decoder with an upstream conditioning encoder and a downstream PET-domain adapter. A two-phase training strategy was used, with the first phase learning coarse uptake distributions and the second refining local image details. Uniform organ activity maps were generated from CT-based segmentations by assigning each organ its mean uptake from the paired PET image. Evaluation included quantitative accuracy, noise assessment, radiomic analysis, tumor segmentation performance, and a human observer study. PAD-generated images achieved high quantitative accuracy, with concordance correlation coefficients above 0.92 between organ mean SUVs and assigned activity values. The synthesized images showed noise levels and texture characteristics similar to target PET images and produced comparable tumor segmentation performance. In a two-alternative forced-choice observer study, four readers achieved approximately 50% accuracy, indicating visual indistinguishability between synthesized and target images. PAD also generated realistic PET images from XCAT-derived activity maps, demonstrating compatibility with phantom-based anatomical priors. Overall, PAD provides a diffusion-based framework for generating clinically relevant heterogeneous PET images from uniform organ activity maps derived from clinical segmentations or digital phantoms, supporting data augmentation and downstream imaging studies.*

## 1. Introduction

Synthetic positron emission tomography (PET) imaging data are of substantial value in the field of PET imaging because they serve as a practical ground truth for developing, optimizing, and validating imaging methods[1,2]. Conventionally, the imaging subject is represented using digital anthropomorphic phantoms, for example the 4D extended cardiac-torso phantom (XCAT)[3], and the PET acquisition process is modeled using physics-based simulation tools, such as the Geant4 Application for Tomographic Emission (GATE)[4]. Together, these components form simulation pipelines for generating synthetic PET images. Their reproducibility and high degree of control over scanner and acquisition settings make them well-suited for applications such as scanner design and the optimization of imaging and reconstruction protocols[5-7]. However, the GATE simulation easily takes several hours to complete for one patient[8], and the resulting images do not fully capture the complex uptake heterogeneity and realistic visual appearance observed in clinical PET images. Moreover, the digital phantom design offers only limited flexibility for customizing patient anatomy. As a result, simulated images do not faithfully represent real image characteristics, making them less suitable for downstream applications that are sensitive to subtle image features such as segmentation and harmonization. This limitation motivates the development of methods to generate more realistic synthetic PET images. Additionally, deep learning has been widely applied in medical image analysis[9,10] but the availability of sufficiently large clinical PET datasets remains limited. Realistic PET image generation therefore has become an important strategy for data augmentation to further empower deep learning-based model development and validation, as well as to support the growth of virtual imaging trials.

Recent years have witnessed wide applications of deep learning techniques to image synthesis. Variational autoencoders (VAEs)[11] are among the earliest deep generative models developed for scalable image generation. However, images generated by VAEs are often overly smooth and lack fine structural detail. Generative adversarial networks (GANs)[12] subsequently advanced the field by introducing adversarial training to enable sharper and more realistic image generation. In PET imaging, GAN-based methods have found broad application in tasks such as parametric image generation[13], modality translation[14], and image denoising[15]. Nevertheless, GANs are often associated with training instability and mode collapse, which can limit the robustness and diversity of the generated results[16]. As such, denoising diffusion probabilistic models (DDPMs)[17] were proposed as an alternative. In DDPMs, noise is gradually added to the data through a forward Markov diffusion process. A neural network is trained to iteratively reverse this process to generate samples from noise, rather than predicting the image directly in a single step. Prior studies have proved that diffusion models can outperform GANs in both natural and medical image synthesis by offering more stable optimization and improved performance in terms of image fidelity and sample diversity[18-20]. These advantages have driven the growing usage of DDPMs in various PET imaging applications[21,22].

In this work, we developed and validated a novel pretrained domain-adapted diffusion (PAD) model to generate realistic clinical PET images from uniform organ activity maps, thereby enabling anatomy-conditioned PET image synthesis. To address data limitations and improve image quality, PAD adopts a two-phase modeling strategy to break the image generation process into coarse distribution prediction and local feature refinement. Additionally, to stabilize training and improve data efficiency, we leveraged a text-to-image model pretrained on natural images to initially anchor the encoder and adapter within its rich semantic space. Finally, for quality control, five validation experiments were conducted to assess different aspects of the synthetic images: quantitative accuracy, noise level similarity, radiomic analysis, task-based validation, and human observer studies. These evaluations demonstrated that PAD produced images with high quantitative accuracy, realistic noise and texture properties, and tumor segmentation performance most consistent with the target images. Moreover, the approximately 50% reader accuracy in the human observer study suggested that PAD-generated and target images were visually indistinguishable. Lastly, PAD's successful application to XCAT-derived inputs further demonstrated its compatibility with phantom-based anatomical priors.

## 2. Methods

### *2.1. Dataset preparation and preprocessing*

#### 2.1.1. Uniform organ activity map generation

A total of 513 FDG-PET/CT images without PET-positive lesions from a publicly available clinical dataset[23] were used as source data to extract patient anatomy and then generate uniform organ activity maps. For each case, the CT image was automatically segmented using TotalSegmentator[24], yielding 128 anatomical structures. Based on these segmentations, uniform activity maps were generated by assigning the mean uptake value from the paired PET image to each masked structure. During training, these uniform activity maps were used as network conditioning inputs, and the corresponding clinical PET images served as target images. Finally, all input maps and target images were normalized according to Section 2.1.2 and cropped to a size of 256 × 256 to reduce blank margins while preserving the patient body within the image.

#### 2.1.2. Image normalization

Normalization is a critical step in the application of deep learning techniques to PET images because of the high variability in tracer uptake across patients and the long-tailed intensity distribution within each PET image. In this study, all PET images underwent a two-step normalization procedure. First, images were converted to standardized uptake value (SUV) to correct for variations in patient body weight and injected dose and thus reduce inter-patient variability. Second, to deal with the high dynamic range and long-tailed intensity distribution characteristic of PET images within each patient, an inverse hyperbolic sine (arcsinh) transformation was applied, as defined in Equation *1*.

$$x_{norm} = arcsinh(x/c) \qquad 1$$

Here, $x$ denotes the original image in SUV units, and $c$ is set to 0.76 as the transformation threshold to preserve an approximately linear response for activity values within regions of interest while compressing extremely high intensities, such as those in the bladder, into the logarithmic regime. Finally, the arcsinh-transformed images were normalized to a range of 0 to 1 using min-max normalization. This normalization compresses outlier intensities while preserving near-zero background values, thereby improving numerical stability and promoting consistent gradient behavior during the diffusion-based generative process.

### *2.2. Pretrained domain-adapted diffusion (PAD) model*

#### 2.2.1. Network training strategy

Generating high-resolution PET images with a diffusion model requires substantial training data, so directly synthesizing full-resolution PET images is challenging with the current dataset. Therefore, we adopted a two-phase coarse-to-fine training strategy inspired by PITI[25]. In the first phase, a base conditional diffusion model was trained to synthesize low-resolution target PET images (size $64 \times 64$, obtained by an area-based downsampling method[26]) conditioned on the corresponding uniform organ activity maps. At this stage, the network learned the global uptake distribution and coarse anatomical structure. In the second phase, a conditional super-resolution diffusion model was trained to reconstruct the final full-resolution PET images from the corresponding downsampled PET images, with the same uniform activity maps providing anatomical and uptake conditioning. This design enabled the model to preserve global anatomical and uptake information from the first phase while refining local image details at full resolution.

In both phases, image generation is formulated as a DDPM model[17]. Given a clean target image $\boldsymbol{x}_0$ for each phase, the forward diffusion process progressively corrupts it by adding independent Gaussian noise over $T$ timesteps ($T = 1000$ in this study). The transition at step $t$ is defined according to Equation *2*:

$$q(\boldsymbol{x}_t \mid \boldsymbol{x}_{t-1}) = \mathcal{N}(\sqrt{1-\beta_t}\boldsymbol{x}_{t-1}, \beta_t \mathbf{I}) \qquad 2$$

where $\boldsymbol{x}_t$ denotes the noisy image at step $t$, $\beta_t$ controls the noise level injected at that step. By the chain rule for a Markov process, the marginal distribution of $\boldsymbol{x}_t$ conditioned on $\boldsymbol{x}_0$ has the closed form

$$q(\boldsymbol{x}_t \mid \boldsymbol{x}_0) = \mathcal{N}(\sqrt{\bar{\alpha}_t}\boldsymbol{x}_0, (1-\bar{\alpha}_t)\mathbf{I}) \tag{3}$$

such that:

$$\boldsymbol{x}_t = \sqrt{\bar{\alpha}_t}\boldsymbol{x}_0 + \sqrt{1-\bar{\alpha}_t}\boldsymbol{\epsilon}, \qquad \boldsymbol{\epsilon} \sim \mathcal{N}(\mathbf{0}, \mathbf{I}) \tag{4}$$

where $\alpha_t = 1 - \beta_t$ and $\bar{\alpha}_t = \prod_{s=1}^{t} \alpha_s$.

At each diffusion timestep $t$, the network takes in the noisy image $\boldsymbol{x}_t$ together with the conditioning input $\boldsymbol{c}$, denoting the uniform organ activity maps, to predict the added Gaussian noise $\boldsymbol{\epsilon}_{\boldsymbol{\theta}}(\boldsymbol{x}_t, t, \boldsymbol{c})$ and the reverse-process variance $\boldsymbol{\Sigma}_{\boldsymbol{\theta}}(\boldsymbol{x}_t, t, \boldsymbol{c})$. The training objective is defined as in Equation *5*.

$$\mathcal{L} = \mathbb{E}_{\boldsymbol{x}_0,\boldsymbol{c},\boldsymbol{\epsilon},t}[\| \boldsymbol{\epsilon} - \boldsymbol{\epsilon}_{\boldsymbol{\theta}}(\boldsymbol{x}_t, t, \boldsymbol{c}) \|_2^2 + \lambda_{vb} D_{KL}(q(\boldsymbol{x}_{t-1} \mid \boldsymbol{x}_t, \boldsymbol{x}_0) \parallel p_{\boldsymbol{\theta}}(\boldsymbol{x}_{t-1} | \boldsymbol{x}_t, \boldsymbol{c})) + \lambda_{L1} L_1(\widehat{\boldsymbol{x}}_0, \boldsymbol{x}_0)] \tag{5}$$

where $\lambda_{vb}$ and $\lambda_{L1}$ are scalar weighting factors introduced to balance the relative contributions of the three loss components. The first term computes the mean squared error between the true Gaussian noise $\boldsymbol{\epsilon}$ and the noise predicted by the network $\boldsymbol{\epsilon}_{\boldsymbol{\theta}}(\boldsymbol{x}_t, t, \boldsymbol{c})$. It trains the network to recover the Gaussian noise added in the forward process and thereby learns the denoising direction required for reverse generation. The second term is the variational bound term that primarily supervises the learned reverse-process variance $\boldsymbol{\Sigma}_{\boldsymbol{\theta}}(\boldsymbol{x}_t, t, \boldsymbol{c})$ by matching the parameterized reverse transition $p_\theta$ to the true posterior $q$. Since the reverse process mean $\boldsymbol{\mu}_{\boldsymbol{\theta}}(\boldsymbol{x}_t, t, \boldsymbol{c})$ is analytically determined by the predicted noise $\boldsymbol{\epsilon}_{\boldsymbol{\theta}}(\boldsymbol{x}_t, t, \boldsymbol{c})$[17]:

$$\boldsymbol{\mu}_{\boldsymbol{\theta}}(\boldsymbol{x}_t, t, \boldsymbol{c}) = \frac{1}{\sqrt{\alpha_t}}\left(\boldsymbol{x}_t - \frac{\beta_t}{\sqrt{1-\bar{\alpha}_t}}\boldsymbol{\epsilon}_{\boldsymbol{\theta}}(\boldsymbol{x}_t, t, \boldsymbol{c})\right) \tag{6}$$

A stop-gradient (sg) operator is applied to $\boldsymbol{\mu}_{\boldsymbol{\theta}}$ when evaluating the variational bound term so that gradients for the variance prediction do not affect the mean prediction branch:

$$p_{\boldsymbol{\theta}}(\boldsymbol{x}_{t-1} | \boldsymbol{x}_t, \boldsymbol{c}) = \mathcal{N}(sg(\boldsymbol{\mu}_{\boldsymbol{\theta}}(\boldsymbol{x}_t, t, \boldsymbol{c})), \boldsymbol{\Sigma}_{\boldsymbol{\theta}}(\boldsymbol{x}_t, t, \boldsymbol{c})) \tag{7}$$

The third term is an image-domain reconstruction loss that penalizes the $L_1$ distance between the predicted clean image $\widehat{\boldsymbol{x}}_0$ and the target image $\boldsymbol{x}_0$, thereby encouraging faithful recovery of image intensity and structural details. Fortunately, the predicted clean image does not need to be predicted independently from the network and can instead be expressed in terms of the network predicted Gaussian noise $\boldsymbol{\epsilon}_{\boldsymbol{\theta}}(\boldsymbol{x}_t, t, \boldsymbol{c})$ by rearranging Equation *4*:

$$\widehat{\boldsymbol{x}}_0 = \frac{\boldsymbol{x}_t - \sqrt{1-\bar{\alpha}_t}\,\boldsymbol{\epsilon}_{\boldsymbol{\theta}}(\boldsymbol{x}_t, t, \boldsymbol{c})}{\sqrt{\bar{\alpha}_t}} \tag{8}$$

### 2.2.2. Network Architecture

The proposed PAD model is built upon a pretrained text-to-image diffusion backbone[27] to fully leverage the robust generative priors embedded in large-scale foundation models. As illustrated in Figure 1, the network consists of three primary modules: a local–global reference encoder, a pretrained diffusion decoder, and a light-weight domain adapter.

The local–global encoder takes reference images, which are the uniform organ activity maps in this study, to extract multi-scale contextual features. A Convolutional Neural Network (CNN) encoder[28] first captures localized spatial details, which are then processed by a Transformer encoder[29] to model long-range global dependencies. It outputs a sequence of semantic conditioning tokens, denoted as $Ref_{enc}$.

The pretrained diffusion decoder is a U-Net architecture initialized with pretrained text-to-image weights from GLIDE[27]. To condition the denoising process on the reference image, the original text-based cross-attention layers located within the U-Net's encoder, bottleneck, and decoder blocks are replaced with cross-attention to the semantic conditioning tokens $Ref_{enc}$. The input to this decoder varies slightly depending on different training phases: for the first-phase base model, the network takes only the noisy sample $\boldsymbol{x}_t$ as its spatial input. For the second-phase super-resolution model, the low-resolution target images are concatenated with $\boldsymbol{x}_t$ to provide explicit spatial guidance and thus focus the model on high-frequency detail synthesis.

Finally, the light-weight domain adapter containing two convolutional layers is appended after the pretrained diffusion decoder. It bridges the distribution gap between the pretrained natural image domain and the target PET image domain.

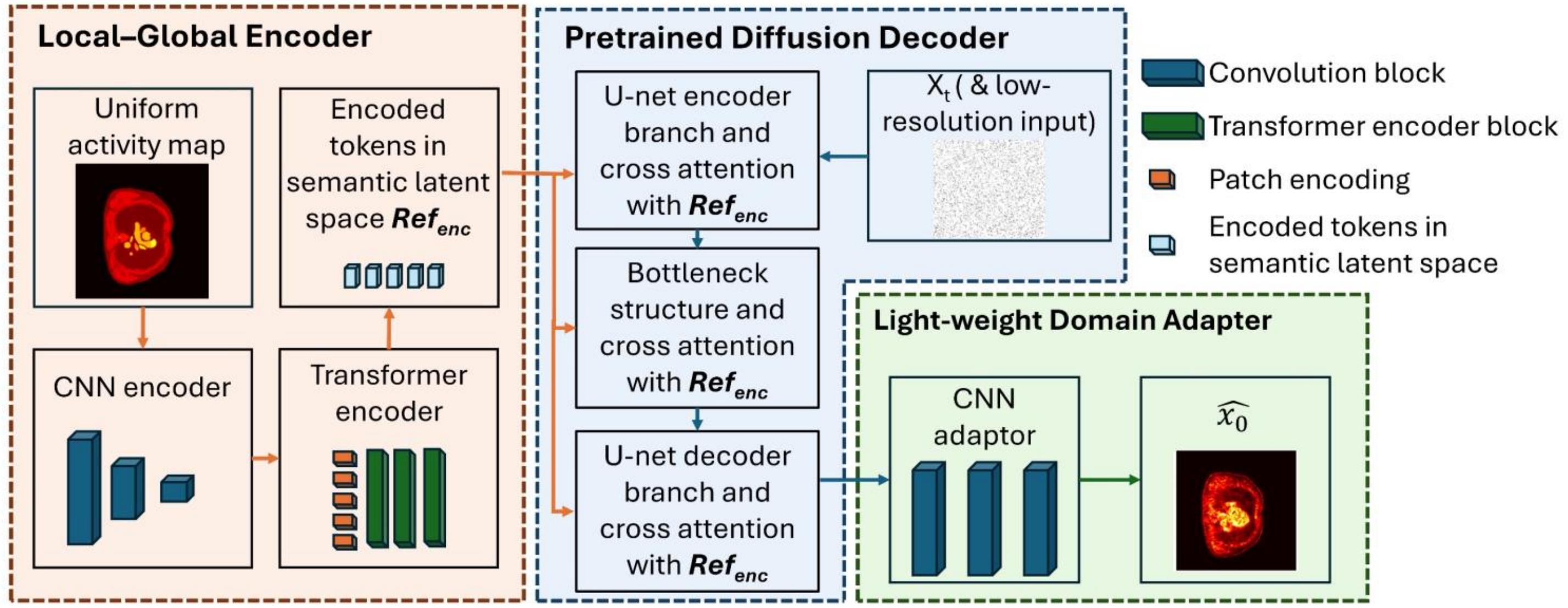

*Figure 1. Overview of the architecture of the pretrained domain-adapted diffusion (PAD) model.*

### 2.2.3. Implementation details

The first-phase model was trained for 460k iterations with an initial learning rate of 1e-5, whereas the second-phase model was trained for 160k iterations with an initial learning rate of 3e-5. In each phase, the pretrained diffusion decoder was initially frozen for the first 160k and 100k iterations, respectively, allowing the encoder and adapter to adapt to the decoder's latent space before all components were jointly fine-tuned. Optimization was performed using the AdamW optimizer[30] with the learning rate linearly decayed to zero. For the forward diffusion process, the first-phase model utilized a squared-cosine noise schedule[31], which provides a smoother progression of noise levels across timesteps, while the second-phase super-resolution model employed the traditional linear noise schedule[17]. Additionally, to optimize the training process and focus optimization on more challenging diffusion steps, we adopted a loss-aware importance sampling strategy[31] that adaptively prioritizes diffusion timesteps according to the historical second moment of training losses. The variational bound weight $\lambda_{vb}$ was set to 1 for both phases, while the $L_1$ reconstruction weight $\lambda_{L1}$ was set to 0.03 in the first phase and 0.02 in the second phase to facilitate model convergence. Finally, an exponential moving average[32] (EMA) of the model parameters, with a decay rate of 0.9999, was maintained throughout training to ensure more stable, high-fidelity generation during inference.

The network was trained in a 2D slice-wise configuration. The dataset was divided into training, validation, and test subsets at an approximate ratio of 3:1:1. Under this split, the training set comprised 313 3D input-target image pairs (approximately 71K 2D slice pairs), whereas the validation and test sets each included 100 pairs (approximately 23.6K 2D slice pairs). The model was implemented using the PyTorch framework version 2.8.0. The training was performed on two NVIDIA RTX 6000 ADA GPUs with CUDA 12.2 support. All experiments were executed in 32-bit precision within a Linux environment running on an x86_64 architecture.

## *2.3. Evaluation*

To comprehensively evaluate the performance of the PAD model, we compared it against two established baseline architectures: pix2pix[33], and ResViT[34]. Pix2pix is a classical CNN-based network in image-to-image translation task and ResViT serves as a state-of-the-art transformer-based model specifically designed for medical image generation. All the images were denormalized to the scale of SUV and evaluated in five different aspects: quantitative value accuracy, noise level similarity, radiomic analysis, tumor segmentation task-based performance comparison, and human observer study.

### 2.3.1. Quantitative accuracy

Given that the objective is to generate PET images from uniform organ activity maps, the generated images are expected to preserve organ-level mean activity values consistent with the assigned inputs. Therefore, quantitative accuracy was evaluated across the test population by comparing the mean SUV values of major organs between the generated and target images, including the lung, heart, liver, vertebrae, and large intestine. Organ masks were derived from the same segmentation masks used to construct the uniform activity maps in Section 2.1.1 . Linear regression and Lin's concordance correlation coefficient (CCC)[35] were used to evaluate the agreement. Finally, two-sided paired bootstrap statistical tests[36] with 1,000 resamples were performed to determine whether the CCC obtained from PAD-generated images differed significantly from those of the other methods.

### 2.3.2. Noise level similarity

The coefficient of variation (CoV)[37,38] is a widely used metric for quantifying noise levels in PET images. Because the liver is a relatively large and homogeneous organ in FDG-PET imaging, the CoV measured within the liver was used in this study as a marker of noise level. For each network, the distribution of liver CoV values across the test set was visualized as a histogram and plotted against that of target images for direct comparison. The histogram binning followed the Freedman-Diaconis rule[39].

### 2.3.3. Radiomic analysis for Quality Assurance

Radiomic features derived from PET images capture clinically relevant information and have been widely associated with diagnosis[40] and prognosis[41]. It is therefore important to evaluate whether these features are preserved in the generated images relative to the target images. In this study, ten first-order features extracted from the liver, lung, and kidney were considered: 10th percentile, 90th percentile, entropy, interquartile range, kurtosis, mean absolute deviation, median, range, skewness, and uniformity. Seventeen grey-level co-occurrence matrix (GLCM) texture features were also extracted from the same organs: autocorrelation, contrast, correlation, difference average, difference entropy, difference variance, inverse difference, inverse difference moment, informational measure of correlation (IMC) 1, IMC 2, inverse variance, joint average, joint energy, joint entropy, maximal correlation coefficient (MCC), sum average, and sum entropy. In GLCM feature calculation, all features were computed along four directions $(0°, 45°, 90°, \text{and } 135°)$ and feature values were averaged across all directions to obtain the final measurements. Finally, for each radiomic feature, the similarity between the population-level distributions derived from the generated and target images in the test set was quantified using the Jensen–Shannon (JS) divergence[42].

### 2.3.4. Tumor segmentation task-based validation

Task-driven evaluation is an essential component in assessing new imaging pipelines[43]. A tumor segmentation task was conducted in this study because of its strong clinical significance and its widespread usage in PET image analysis[44,45]. Specifically, both synthesized and target images were treated as lesion-free background images. For each patient in the test set, the same simulated tumor was inserted at the same location in both the synthesized and corresponding target images. A deep learning–based segmentation network was then applied to delineate the inserted tumors. Lastly, the difference in segmentation performance between synthesized and target images was compared across generation methods.

The inserted tumors were designed to mimic realistic lung lesions. Tumor size, shape, and signal-to-background ratio were randomly sampled from lung tumors in the FDG-PET-CT-Lesions dataset[23]. Intratumoral heterogeneity was modeled using the Gaussian lumpy model[46,47]. PETSTEP[48] was then used to insert these simulated tumors into randomly selected lung regions in both the synthesized and target images. Tumor segmentation was performed using the first-place model from the AutoPET III challenge[49]. Segmentation performance differences between generated and target images were quantified in terms of the Dice similarity coefficient (DICE)[50] and relative volume difference (RVD)[51] and statistical significance was assessed using a two-sided Wilcoxon signed-rank test[52].

#### 2.3.5. Human observer study

To examine the overall perceived realism of images to humans, a two-alternative forced-choice (2-AFC) reader study[53] was performed with four experienced medical image analysts. All readers were shown a target image and its corresponding PAD–generated image using a web-based software[54] and asked to identify the synthetic image. In total, each reader was presented with 50 unlabeled image pairs and also asked to provide a confidence level along with each decision. The confidence scores were recorded on a 5-point scale, representing 0% to 100% confidence. In a 2-AFC study, observer accuracy provides a measure of perceptual separability. An accuracy of approximately 50% corresponds to an AUC of roughly 0.5 for the equivalent detection task[55] and thus implies that the synthetic images and target images are visually indistinguishable to human readers.

#### 2.3.6. Application to XCAT

To further evaluate the generalizability of PAD, we examined whether the network could generate realistic PET-like images from XCAT-based uniform organ activity maps. In this experiment, XCAT phantoms with 56 different anatomical configurations were used to provide patient-specific anatomical structures. For each XCAT anatomy, organ activity values were assigned from a randomly selected subject in the test population without replacement. The resulting 56 XCAT-based uniform organ activity maps were then provided to PAD for image generation. To assess the realism of the generated images, CoV and five GLCM texture features, including autocorrelation, contrast, difference entropy, difference average, and difference variance, were calculated within the liver region. The distributions of these features were visualized using histograms and compared with those from the target PET images to exhibit the overlap between generated and clinical image feature distributions, thereby assessing the generalizability of the method to XCAT-based inputs.

### *2.4. Statistical analysis*

In the quantitative accuracy validation (Section 2.3.1), the statistical significance of the CCC difference was assessed using a two-sided paired bootstrap test. For the tumor segmentation task, differences in segmentation performance between target and synthesized images were evaluated using a two-sided Wilcoxon signed-rank test. In all statistical tests mentioned above, $p \leq 0.05$ was considered significant, followed by post-hoc pairwise comparisons, which were deemed significant at $p \leq 0.025$ after Bonferroni correction[56]. Furthermore, for the noise level and radiomic feature comparison, JS divergence was used to measure the similarity of all features between synthesized and target images in the test set distributions. All statistical analyses were conducted in Python 3.9.16 using SciPy version 1.13.1.

## 3. Results

To provide a visual comparison, Figure 2 displays representative slices of target images, uniform organ activity maps, and synthesized images from various networks (in SUV units).

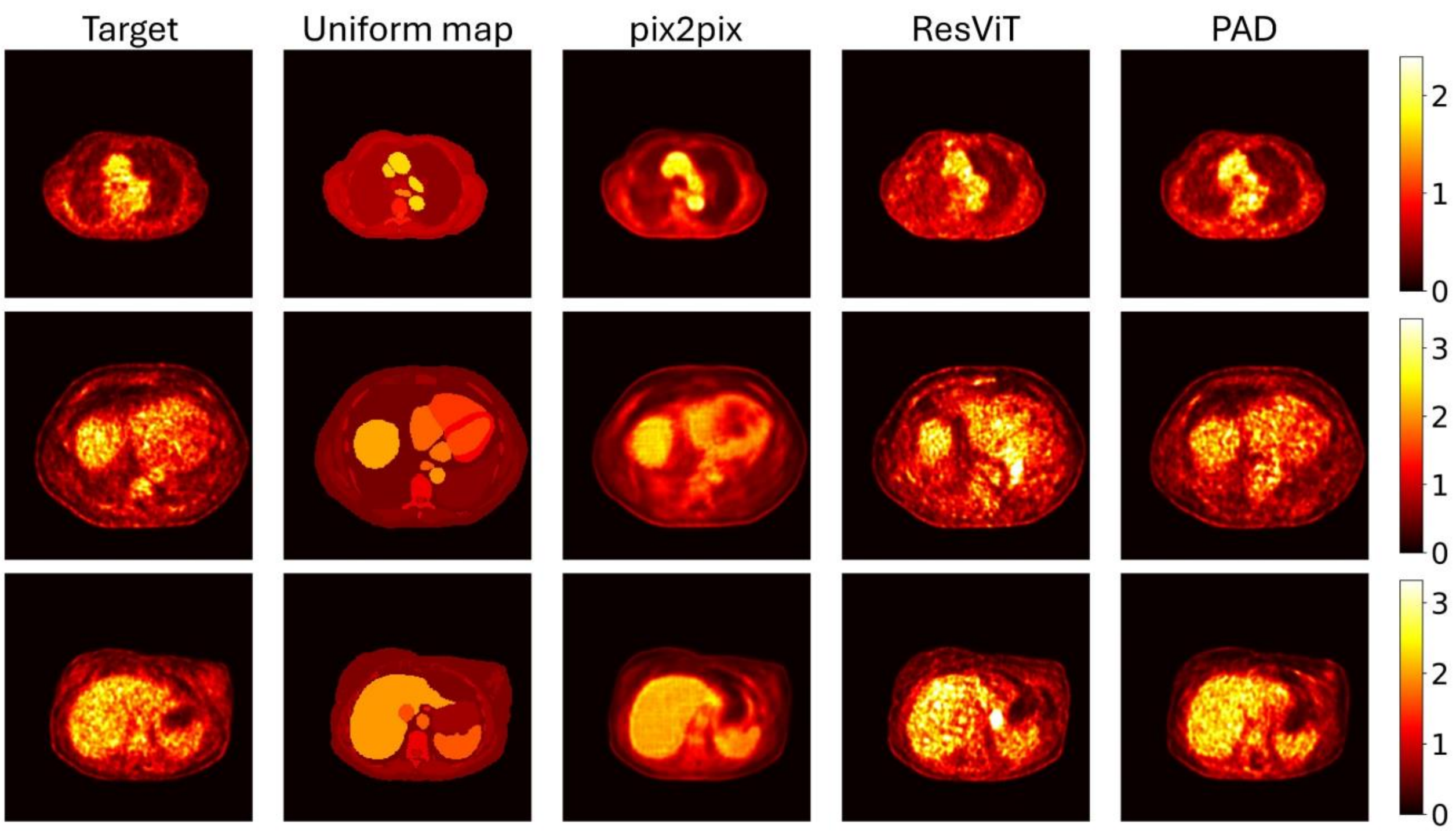

*Figure 2. Visual comparison of target images, uniform organ activity maps, and generated images from various networks (at columns) in SUV units across three transaxial slices (at row).*

## *3.1. Quantitative accuracy*

Figure 3 depicts that the mean SUV values of major organs in the generated images are all positively correlated with the assigned values for all methods. PAD achieved the best overall agreement, with CCC values exceeding 0.92 for all evaluated organs, specifically 0.96 for lung, 0.92 for heart, 0.95 for liver, 0.97 for vertebrae, and 0.92 for large intestine. By comparison, pix2pix and ResViT showed lower agreement, particularly in the lung, heart, and large intestine. Although pix2pix yielded the highest CCC in the liver, PAD was the only method to maintain consistently strong agreement across organs. Furthermore, bootstrap testing confirmed that the CCC values obtained with pix2pix and ResViT were significantly lower than those of PAD for most evaluated organs. These findings indicate that PAD more accurately preserves organ-level quantitative uptake values.

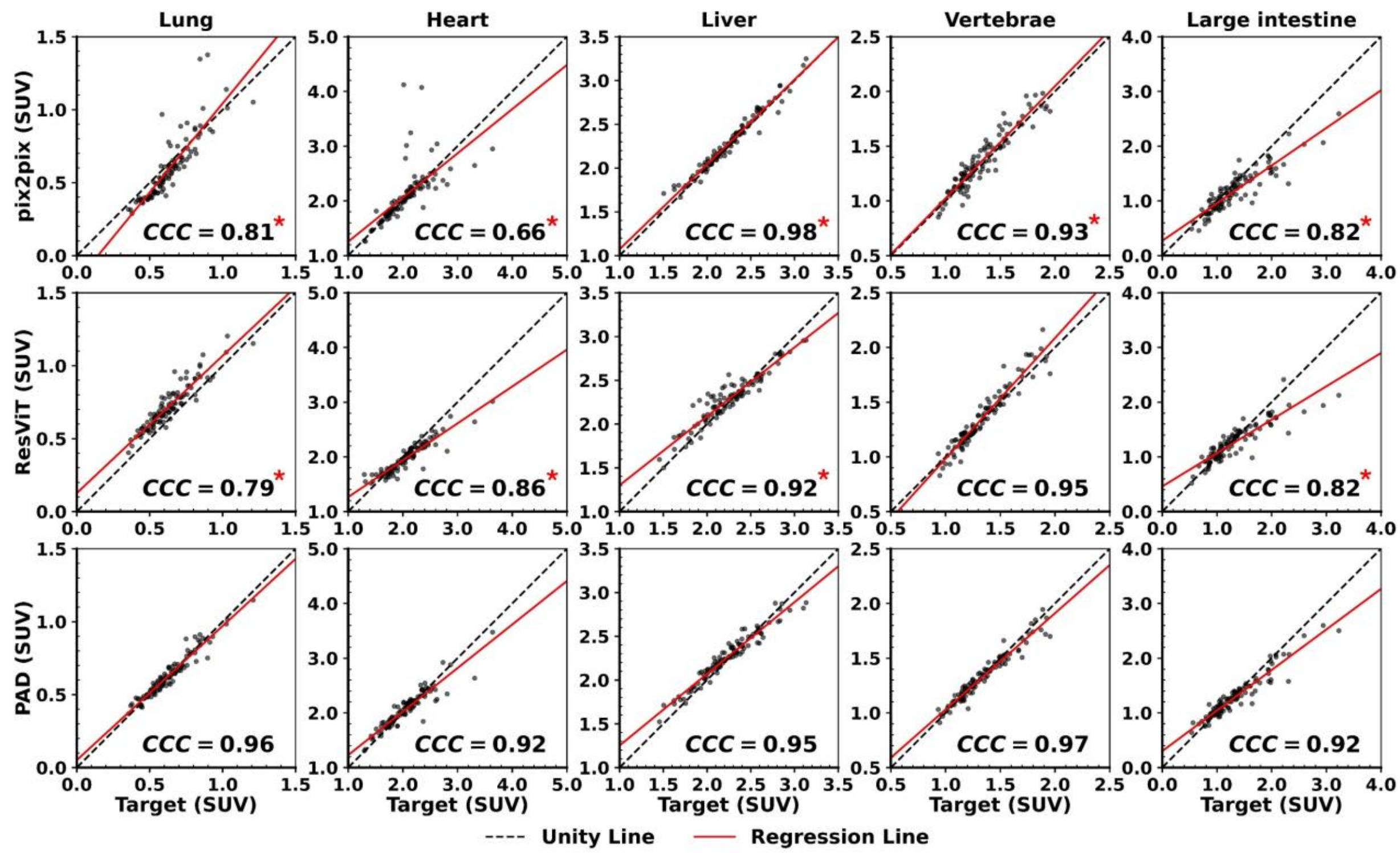

*Figure 3. Linear regression plots along with the CCC values of the mean SUV in major organs between target images and those generated by different networks. The asterisks highlight cases where the CCC values derived from images generated by compared networks are statistically significantly different from PAD in bootstrap tests.*

## 3.2 Noise level similarity

Figure 4 shows clear differences in liver CoV distributions among the evaluated networks. PAD yields the closest agreement with the target images, as reflected by the smallest JS divergence (0.17), whereas ResViT demonstrates moderate agreement (JS = 0.55) and pix2pix exhibits the largest deviation from the target distribution (JS = 0.88). The histogram overlap further indicates that PAD most accurately reproduces the noise characteristics of the target PET images, while pix2pix produces noticeably lower CoV values and ResViT displays a right-shifted and broader distribution relative to the target.

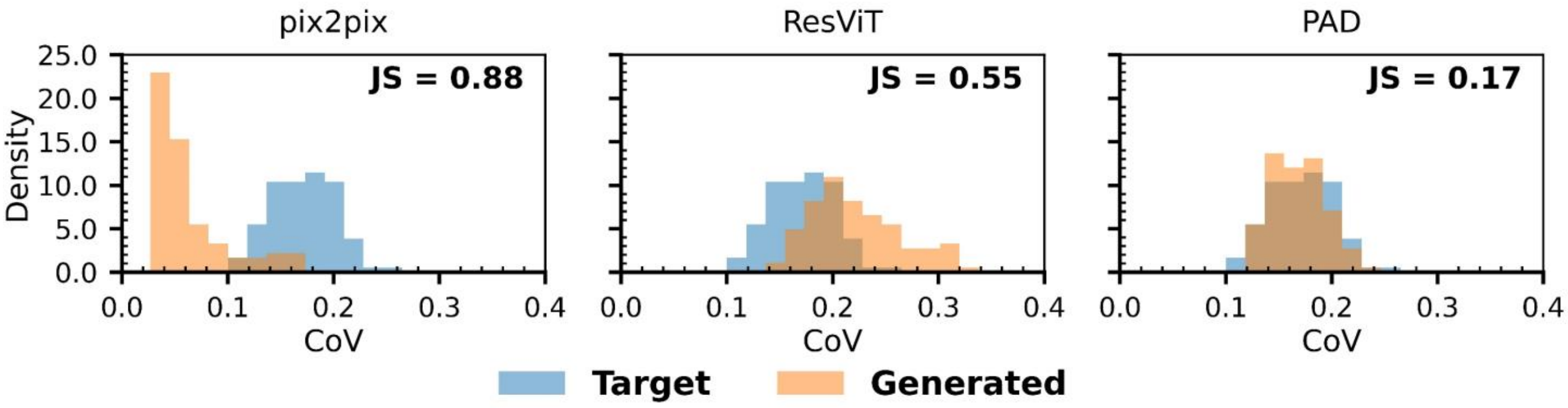

*Figure 4. Distributions of the coefficient of variation (CoV) within the liver across the testing population, comparing target images and synthesized images.*

## 3.3 Radiomic analysis

Table 1. Jensen–Shannon (JS) distance between first-order features within the liver area for target images and images generated by various networks (the lowest JS distance for each feature is shown in bold).Table 1 and Table

2 illustrate the JS distances between the liver radiomic feature distributions of generated and target images. Overall, PAD achieves the closest agreement with the target distributions for most first-order and GLCM features, indicating better preservation of radiomic characteristics. Among the first-order features, PAD produces the lowest JS distance for 8 of 10 metrics, while ResViT performs best for the other two features. For GLCM features, PAD again shows the best agreement for 14 out of 17 metrics while ResViT outperformed PAD for the remaining three. Results for the lung and kidney can be found in Table 1-4 in the supplementary file. Across all feature analyses, pix2pix consistently exhibited the largest divergence from the target distributions. Together, these results suggest that PAD most accurately preserves radiomic feature distributions at the population level.

*Table 1. Jensen–Shannon (JS) distance between first-order features within the liver area for target images and images generated by various networks (the lowest JS distance for each feature is shown in bold).*

| Feature | pix2pix | ResViT | PAD |
|---|---|---|---|
| **10 percentile** | 0.139 | 0.046 | **0.034** |
| **90 percentile** | 0.103 | **0.027** | 0.027 |
| **Entropy** | 0.371 | 0.097 | **0.032** |
| **Interquartile range** | 0.512 | 0.044 | **0.029** |
| **Kurtosis** | 0.506 | 0.245 | **0.076** |
| **Mean absolute deviation** | 0.356 | 0.057 | **0.034** |
| **Median** | 0.044 | **0.031** | 0.039 |
| **Range** | 0.252 | 0.225 | **0.079** |
| **Skewness** | 0.356 | 0.230 | **0.064** |
| **Uniformity** | 0.365 | 0.049 | **0.043** |

*Table 2. Jensen–Shannon (JS) distance between Gray-Level Co-Occurrence Matrix (GLCM) features within the liver area for target images and images generated by various networks (the lowest JS distance for each feature is shown in bold).*

| Feature | pix2pix | ResViT | PAD |
|---|---|---|---|
| **Autocorrelation** | 0.050 | 0.018 | **0.015** |
| **Contrast** | 0.551 | 0.078 | **0.027** |
| **Correlation** | 0.443 | 0.428 | **0.401** |
| **Difference average** | 0.653 | 0.122 | **0.017** |
| **Difference entropy** | 0.595 | 0.091 | **0.011** |
| **Difference variance** | 0.422 | 0.028 | **0.021** |
| **Inverse difference** | 0.673 | 0.162 | **0.018** |
| **Inverse difference moment** | 0.675 | 0.174 | **0.019** |
| **Imc1** | 0.693 | 0.483 | **0.072** |
| **Imc2** | 0.527 | 0.448 | **0.348** |
| **Inverse variance** | 0.395 | 0.160 | **0.021** |
| **Joint average** | 0.041 | 0.018 | **0.012** |
| **Joint energy** | 0.618 | **0.031** | 0.069 |
| **Joint entropy** | 0.506 | **0.022** | 0.042 |
| **MCC** | 0.295 | 0.255 | **0.101** |
| **Sum average** | 0.041 | 0.018 | **0.012** |
| **Sum entropy** | 0.350 | **0.066** | 0.090 |

### 3.4. Tumor segmentation task-based validation

Figure 5 shows that PAD yields the smallest difference in tumor segmentation performance between the generated and target images across all evaluated metrics. Specifically, PAD achieves the lowest mean absolute differences in both DICE and RVD, whereas ResViT displays the largest discrepancies. Statistical analysis further suggests that the performance differences derived from images generated by both pix2pix and ResViT are significantly greater than those from PAD in terms of both DICE and RVD after Bonferroni correction. These results indicate that PAD more faithfully replicates tumor segmentation performance relative to the target images.

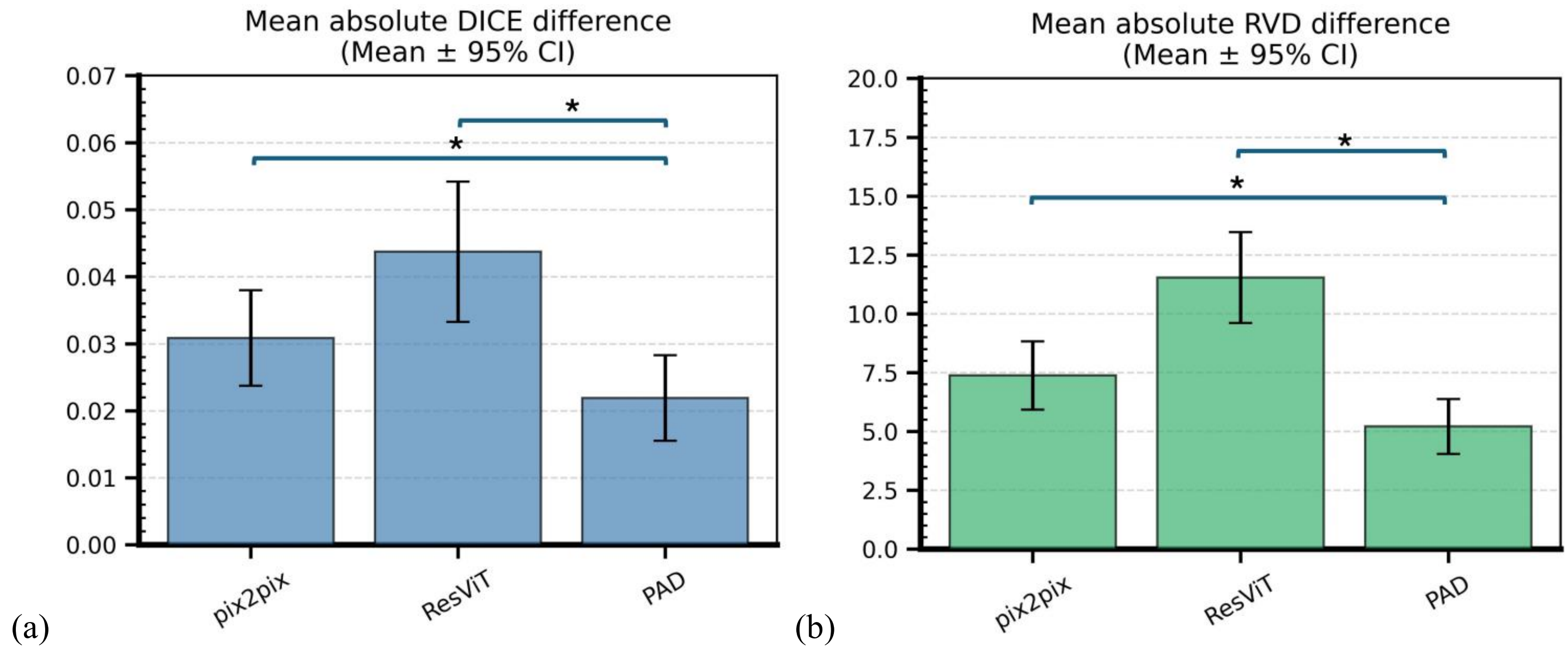

*Figure 5. Bar plots with 95% confidence interval as error bar of tumor segmentation performance difference between target and generated images in terms of (a) Dice similarity coefficient (DICE) and (b) relative volume difference (RVD). The asterisks highlight cases where the difference derived from images generated by compared networks is significantly higher than that from PAD, with 95% confidence after Bonferroni correction in the Wilcoxon signed-rank test.*

### 3.5. Human observer study

Table 3 summarizes the performance of the four readers in the 2-AFC study. Observer accuracy ranges from 42% to 58%, with an overall mean accuracy of 48%, which is close to the chance level. This suggests that the PAD-generated images are largely indistinguishable from the target images to human observers. Confidence levels are also modest, with a median score of 3 for three readers and overall. As a result, these findings indicate that readers are generally unable to reliably differentiate PAD-generated images from real target images, further supporting the visual realism of PAD-synthesized images.

*Table 3. Percent accuracy and median confidence level for each human observer and collectively for all participants in the two-alternative forced choice(2-AFC) study.*

| | **Accuracy** | **Median confidence level** |
|---|---|---|
| Observer 1 | 44% | 3 |
| Observer 2 | 58% | 3 |
| Observer 3 | 48% | 4 |
| Observer 4 | 42% | 3 |
| Summary | 48% | 3 |

### *3.6 Application to XCAT*

Representative slices in Figure 6 demonstrate that PAD successfully generated PET-like images that matched the XCAT anatomy with visually realistic spatial uptake variations and anatomically plausible PET appearance. The CoV and texture feature analysis further support this observation. As shown in Figure 7, the generated images show substantial overlap with the target images in the distributions of liver CoV and five GLCM texture features. This suggests that the network can extend beyond patient-derived anatomical inputs to generate visually realistic PET images from phantom-based anatomical configurations.

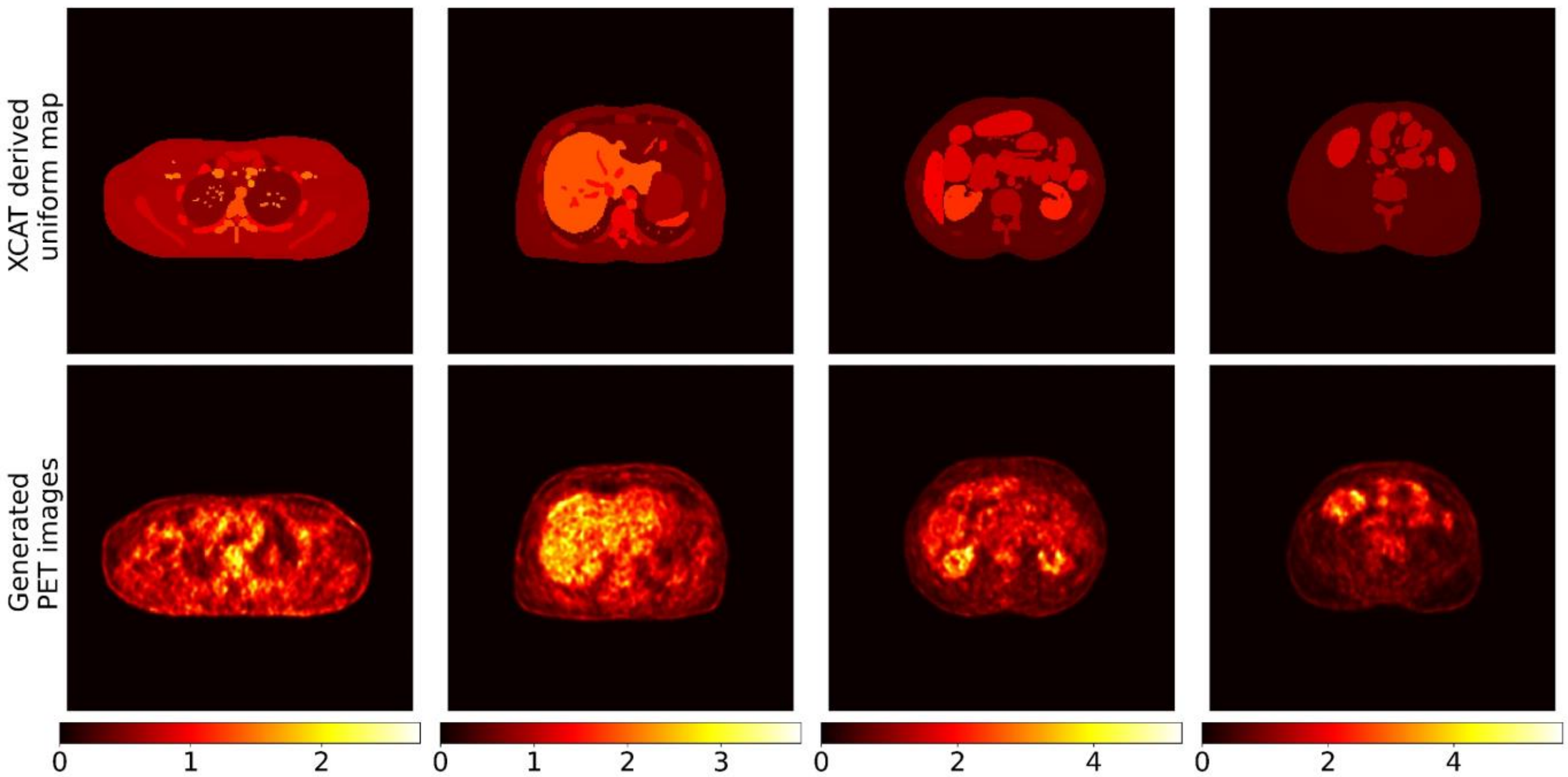

*Figure 6. Representative slices of XCAT-derived uniform organ activity maps (top row) and the corresponding PAD synthesized images (bottom row) in SUV units.*

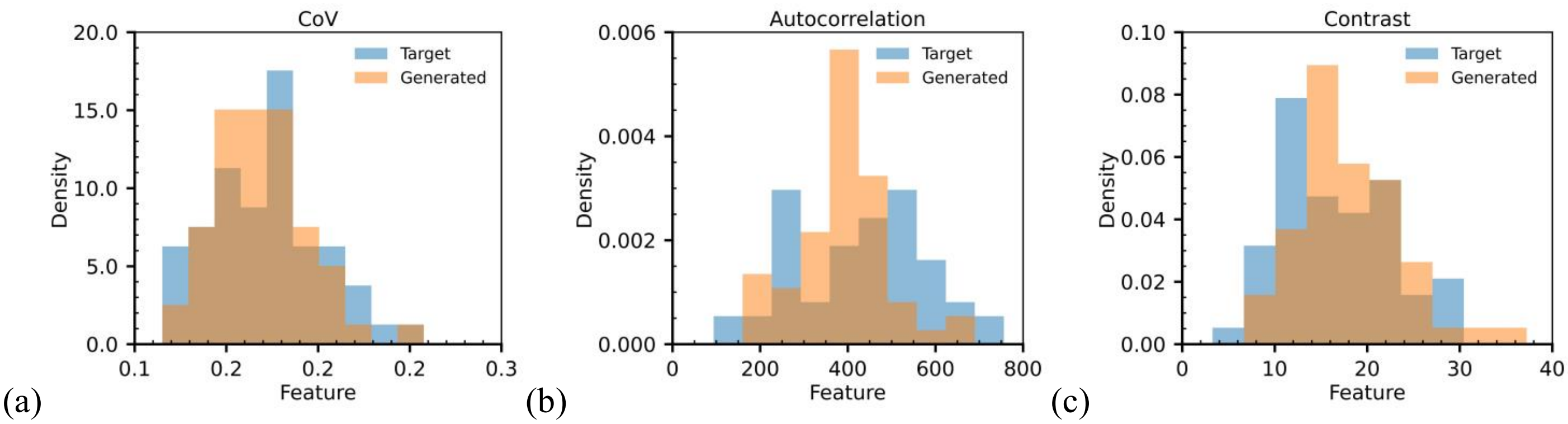

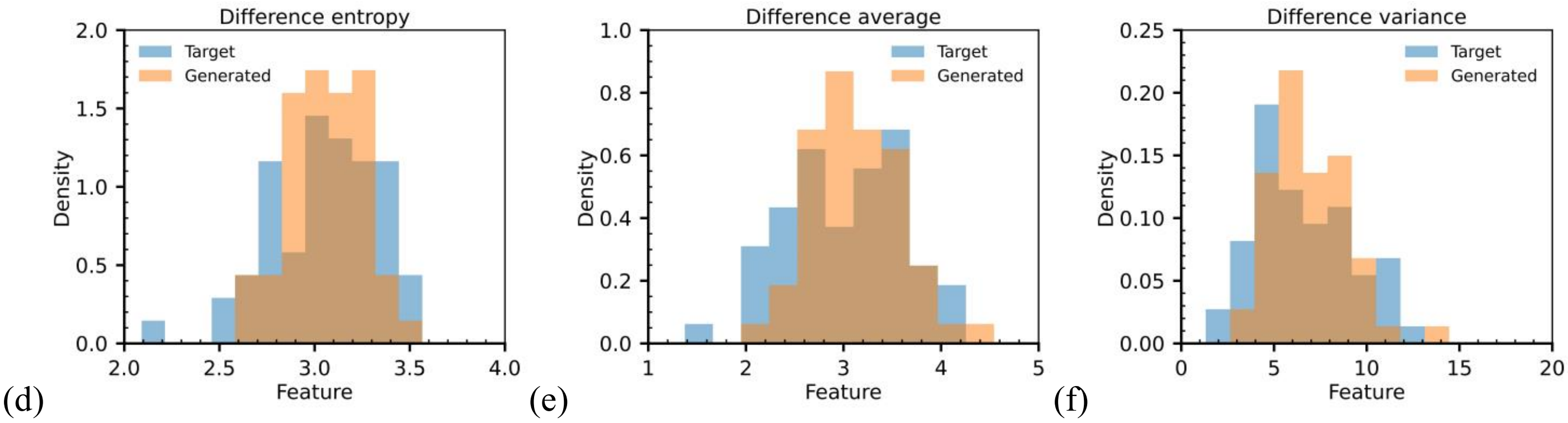

*Figure 7. Histograms comparing the distributions of (a) CoV, (b) autocorrelation, (c) contrast, (d) difference entropy, (e) difference average, and (f) difference variance within the liver area between target PET images and PAD-generated PET images obtained from XCAT-based uniform organ activity maps.*

## 4. Discussion

Synthetic PET images play an important role in task-based image applications, as they enable imaging pipelines and analysis methods to be evaluated under conditions that approximate clinical practice. In this work, we developed PAD, a diffusion–based framework for generating heterogeneous PET images from uniform organ activity maps. By learning from real patient data, PAD generated PET images that remained faithful to the input anatomy while preserving the assigned organ-level activity values and introducing realistic intra-organ uptake heterogeneity. Notably, the input anatomy can be derived either from clinical image segmentations or from digital human phantoms, providing proof of concept that the proposed framework may be applicable across different anatomical priors. PAD also provides a substantially faster approach for PET image synthesis than conventional GATE simulation, which can require hours to complete a single clinical patient simulation, even when using hundreds of CPU cores[38]. In contrast, PAD takes approximately 14 minutes on two NVIDIA RTX 6000 ADA GPUs. A pretrained text-to-image diffusion model and a two-phase training strategy were applied to improve training stability, enhance data efficiency, and facilitate high-quality PET image synthesis. In addition, we conducted comprehensive evaluation experiments to determine whether the generated images were suitable for both quantitative evaluation and clinically relevant tasks. First, the correlation of organ-level mean SUV values was used to evaluate whether the generated images preserved the assigned activity levels. Second, the agreement in CoV and other radiomic features was assessed to characterize image noise properties and textural similarity. Clinical applicability was further examined through a tumor segmentation task. Finally, four human observers completed 2-AFC tests to confirm the texture similarity and visual indistinguishability.

Conditional image synthesis has evolved from GAN-based approaches[57,58] to diffusion-based models[59] in recent years. GANs were first proposed as a model to generate sharp and visually realistic images with relatively efficient sampling. However, GAN-based models are susceptible to training instability and mode collapse, which can limit the diversity and reliability of the generated images. Subsequently, diffusion models were proposed to improve training robustness and preserve sample variability by replacing adversarial training with a more stable denoising objective. Nevertheless, training a diffusion model from scratch is notoriously data-intensive[60]. Hence, we adopted a two-phase generation strategy. The first phase learns the global uptake distribution at low resolution. The second phase refines local details and restores full-resolution image characteristics. This decomposition reduces the difficulty of directly generating high-resolution PET images in a single step and helps balance global consistency with local realism. Moreover, the use of a pretrained model provides the network with a structured latent space and a high-fidelity decoder. It serves as a powerful inductive bias that enables the model to exploit robust feature representations and thereby facilitate model convergence. Within this framework, the encoder projects the conditioning image into the pretrained latent space, while the adapter transforms the decoder output into the PET image domain. Together, these components effectively steer the decoder's prior knowledge toward PET image synthesis, ensuring stable training and high image generation quality.

Virtual imaging trials are gaining increasing attention as a framework to evaluate new imaging systems, reconstruction methods, and downstream analysis pipelines. This approach is not only more cost-effective but also provides a reproducible, ethically sound alternative to traditional clinical studies[61,62]. The proposed method can serve as a tool for generating PET images from user-specified anatomical structures and organ-level activity distributions, thereby enabling controlled virtual imaging of diverse subjects. At the same time, deep learning techniques have been widely employed in PET imaging, with applications in detection[63], segmentation[64,65], and model observer development[66]. These applications require large and robust datasets for effective training and validation. The proposed method helps alleviate the dataset constraints by enabling controllable and realistic PET image generation. Finally, compared with conventional PET image generation pipelines, such as the XCAT–GATE simulation framework, the proposed method introduces realistic PET heterogeneity. This is increasingly important as the development of imaging methods and applications is increasingly expected to maintain clinical relevance, thereby requiring more realistic synthetic PET data for their development and validation.

One limitation of this work is that PAD was trained and validated in a specific imaging context: FDG PET studies acquired on a Biograph mCT scanner under the settings reported in the dataset[23]. Consequently, it does not directly generalize to other tracers, scanner systems, or reconstruction protocols without additional model adaptation. Transfer learning may provide a practical solution to this limitation, as it has been widely used to adapt well-trained models to new datasets with limited additional data and computational cost[67]. Furthermore, PAD is designed with a lightweight adapter that specifically handles domain transformation, making the framework well-suited for transfer learning. To adapt the model to other scanners or imaging domains, fine-tuning can be focused primarily on this adapter. This simplifies the transfer process significantly. Accordingly, an important direction for future work is to expand PAD to a wider range of imaging settings. Moreover, it is also worth noting that PAD did not outperform the comparison methods in every individual metric. For example, pix2pix outperformed PAD in the liver in quantitative accuracy validation (Figure 3), and ResViT showed better similarity than PAD in a few radiomic features (Table 1, Table 2). PAD was considered the best method because it demonstrated the most consistent and superior performance across the full set of evaluations. Nevertheless, these results indicate that further refinement remains important. In addition, because PAD was trained in a 2D slice-based manner while PET is inherently a 3D imaging modality, it may not fully capture cross-slice anatomical continuity. Future work will focus on adapting the network structure to a 3D framework and modifying the training strategy to achieve better global quantitative accuracy and preserve more imaging features.

## 5. Conclusion

We proposed PAD, a diffusion–based framework for anatomy-conditioned PET image synthesis. PAD generates heterogeneous PET images from uniform organ activity maps derived from either clinical segmentations or human phantoms, such as XCAT. The generated images were rigorously evaluated across five dimensions, including quantitative accuracy, noise level similarity, radiomic feature preservation, tumor segmentation task-based performance, and human observer study. It provides a controllable approach for realistic PET image generation and may support a broad range of applications in virtual imaging trials, data augmentation, and the development and validation of task-based imaging methods. Further study is needed to support broader application across multiple domains.